\documentclass[10pt,twocolumn,letterpaper]{article}

\usepackage{cvpr}              % To produce the CAMERA-READY version
\usepackage{etoolbox}
\patchcmd{\thebibliography}
    {\settowidth\labelwidth{[999]}}
    {\settowidth\labelwidth{[999]} \setlength{\itemsep}{0pt}}
    {}
    {}

\usepackage{natbib}
\setcitestyle{numbers,sort&compress,max-names=1}

\definecolor{cvprblue}{rgb}{0.21,0.49,0.74}
\usepackage[pagebackref,breaklinks,colorlinks,allcolors=cvprblue]{hyperref}

\usepackage{multirow}

\usepackage{geometry}
\def\paperID{*****} % *** Enter the Paper ID here
\def\confName{CVPR}
\def\confYear{2025}

\title{Robust sensor fusion against on-vehicle sensor staleness}

\author{Meng Fan, Yifan Zuo, Patrick Blaes, Harley Montgomery, Subhasis Das\\
Zoox Inc\\
1149 Chess Dr, Foster City, CA 94404\\
{\tt\small \{mfan, yzuo, patrick, hmontgomery, subhasis\}@zoox.com}
}

\begin{document}
\maketitle
 
\begin{abstract}
Sensor fusion is crucial for a performant and robust Perception system in autonomous vehicles, but sensor staleness—where data from different sensors arrives with varying delays—poses significant challenges. 
Temporal misalignment between sensor modalities leads to inconsistent object state estimates, severely degrading the quality of trajectory predictions that are critical for safety.
We present a novel and model-agnostic approach to address this problem via (1) a per-point timestamp offset feature (for LiDAR and radar both relative to camera) that enables fine-grained temporal awareness in sensor fusion, and (2) a data augmentation strategy that simulates realistic sensor staleness patterns observed in deployed vehicles. Our method is integrated into a perspective-view detection model that consumes sensor data from multiple LiDARs, radars and cameras. We demonstrate that while a conventional model shows significant regressions when one sensor modality is stale, our approach reaches consistently good performance across both synchronized and stale conditions.
\end{abstract}

\section{Introduction}

Reliable object detection is fundamental to autonomous vehicle (AV) safety, with multi-sensor fusion emerging as a key strategy~\cite{fadadu2022multi,foucard2024spotnet,hwang2022cramnet,bai2022transfusion}. By combining complementary information from LiDAR, camera and radar, AVs can better understand their environment across diverse operating conditions. However, the real-world deployment of such systems faces a critical challenge: sensor staleness, where data from different sensors becomes temporally misaligned due to varying processing delays, system latencies, or hardware constraints~\cite{yuan2022licas3,liu2021matter}.

Although there have been active studies on LiDAR–camera calibration~\cite{rehder2016general,yuan2020rggnet,della2019unified} and synchronization~\cite{yuan2022licas3,liu2021matter}, the sensor staleness problem and its impact on multi-sensor object detection have been less studied. Even when sensor modalities are clock-synchronized and phase-locked at data capture time, processing and transmission delays can cause some modalities to arrive later than others, resulting in stale data.  
When analyzing the temporal characteristics of data feeding into an object detector, we define sensor staleness $t^s$ for a particular sensor modality (e.g., camera, represented by subscript $C$) as:
\begin{equation}
  t^s_C = T_C^{\text{on-time}} - T_C^{\text{current}}
  \label{eq:staleness}
\end{equation}
where $T_C^{\text{on-time}}$ is the ideal camera capture timestamp that should synchronize with the latest data from all sensor modalities received at the moment, and $T_C^{\text{current}}$ is the camera capture timestamp of the most recent camera data actually received.

\begin{figure}[t]
  \includegraphics[scale=0.32]{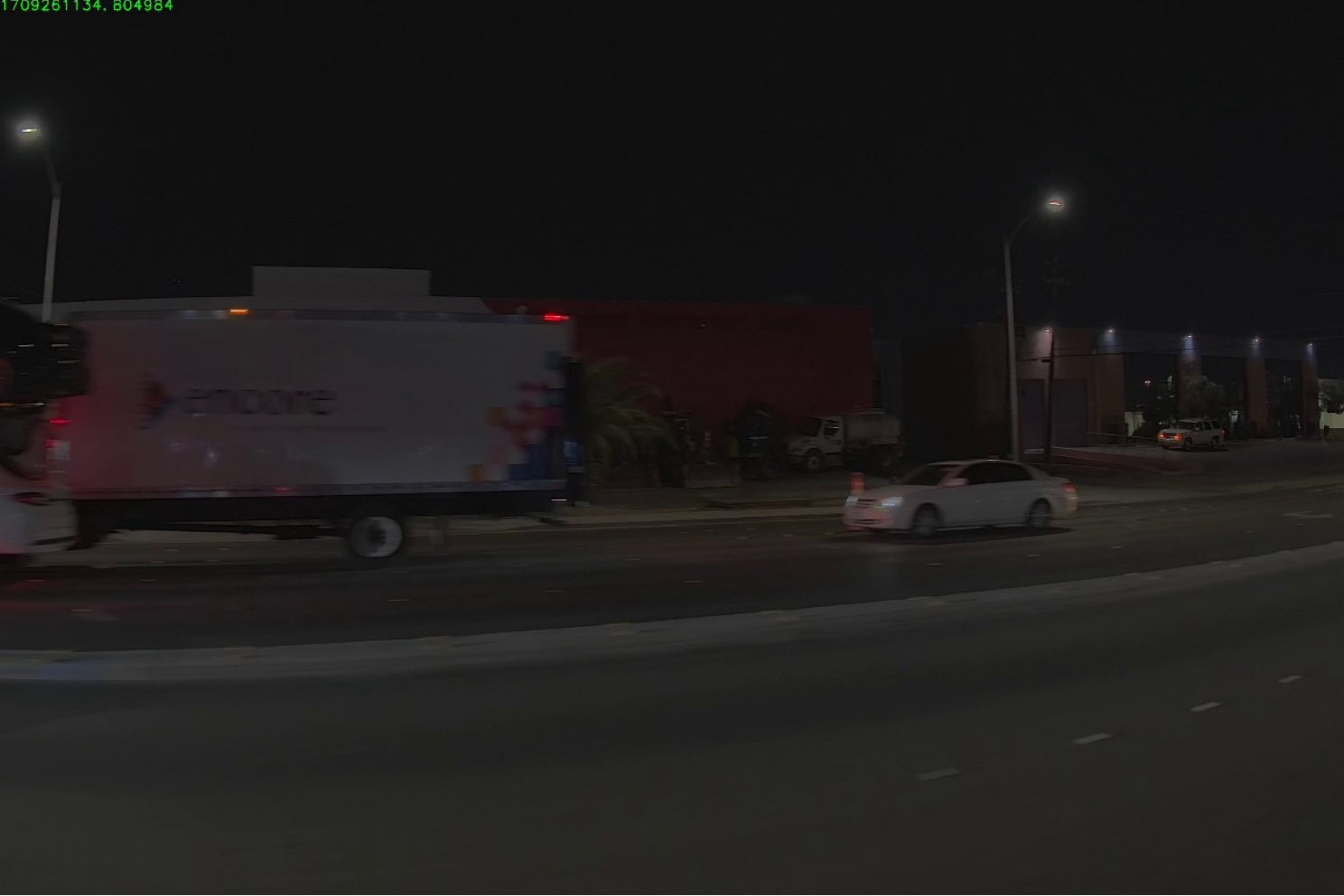}
  \includegraphics[scale=0.32]{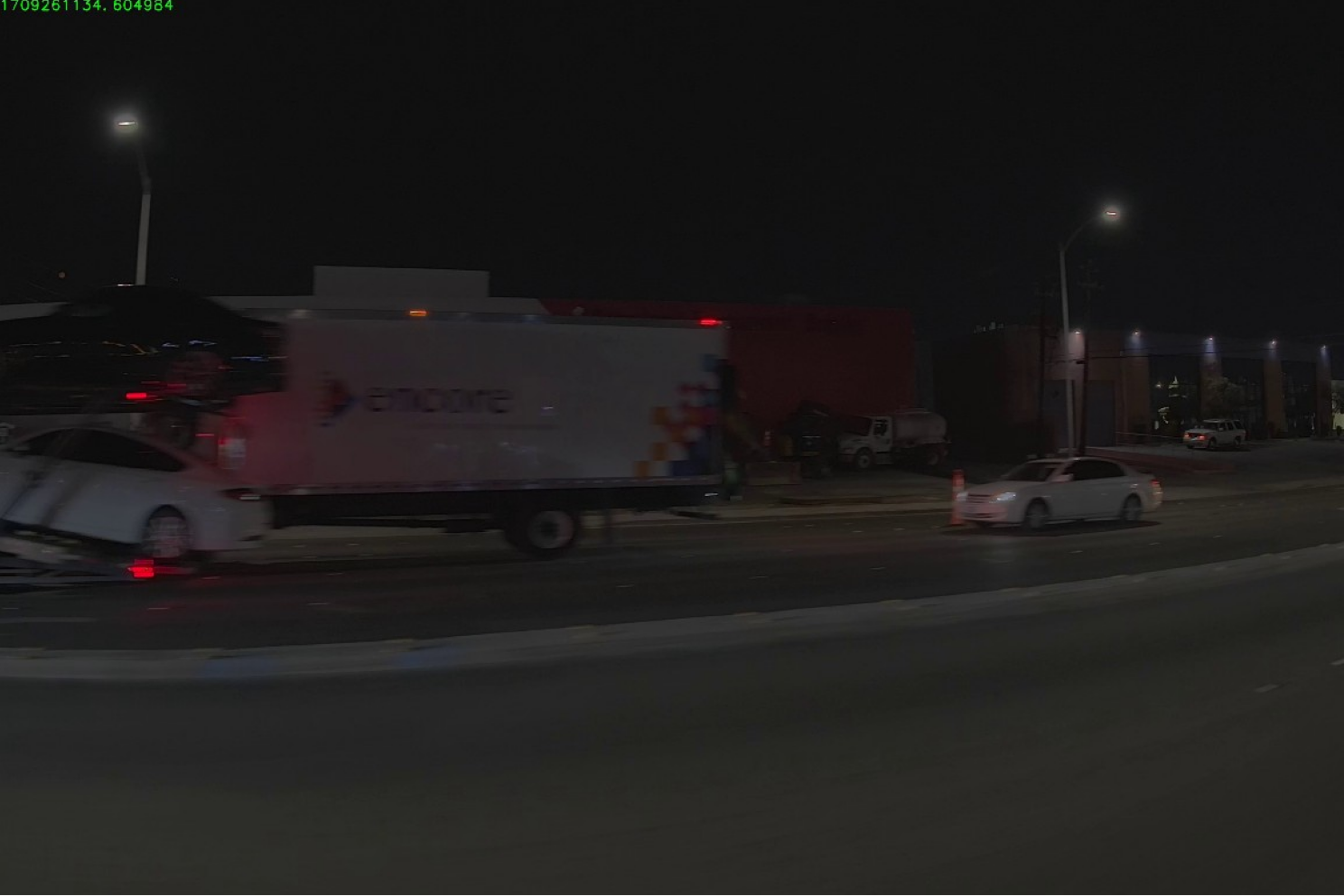}
  \includegraphics[scale=0.32]{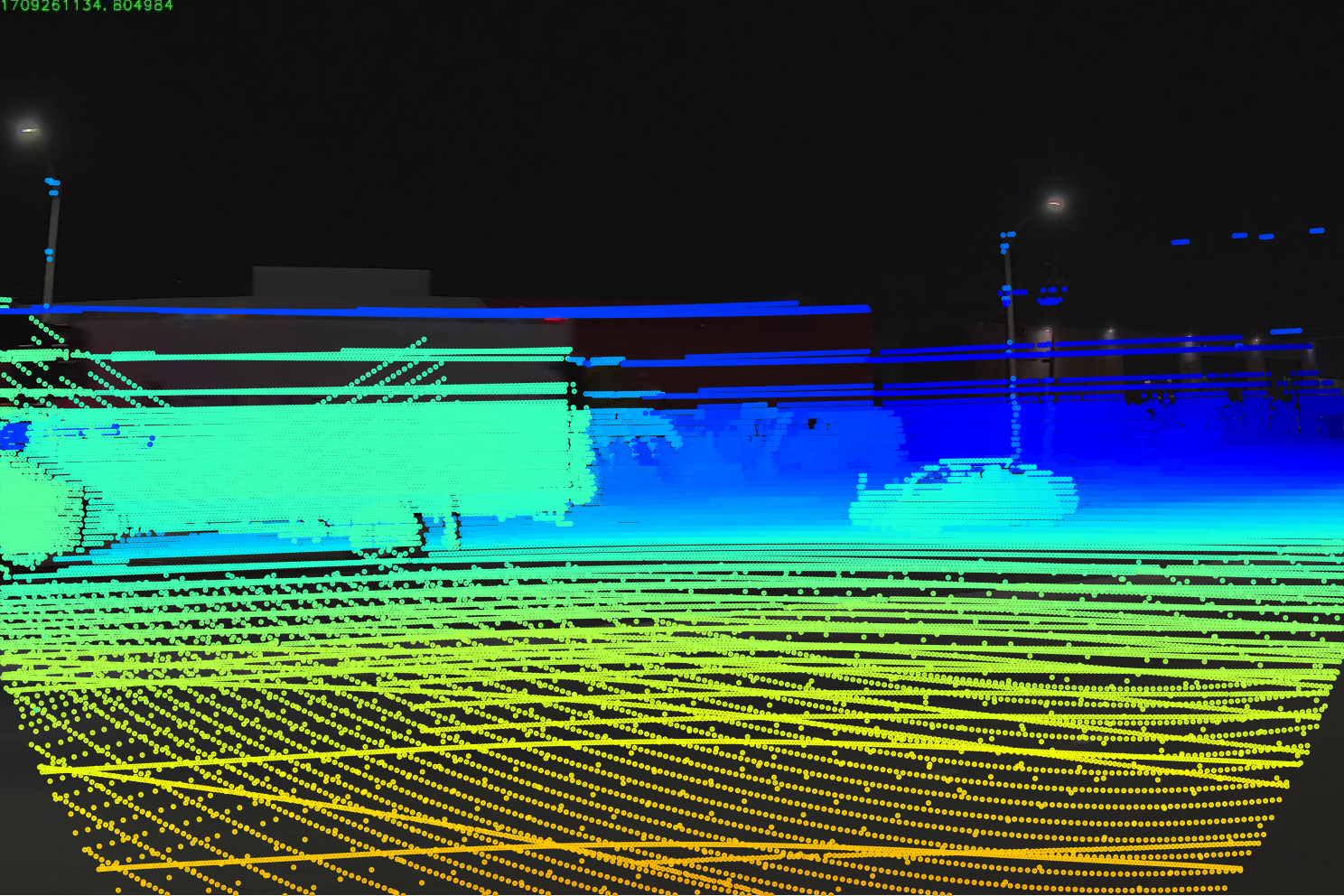}
  \includegraphics[scale=0.32]{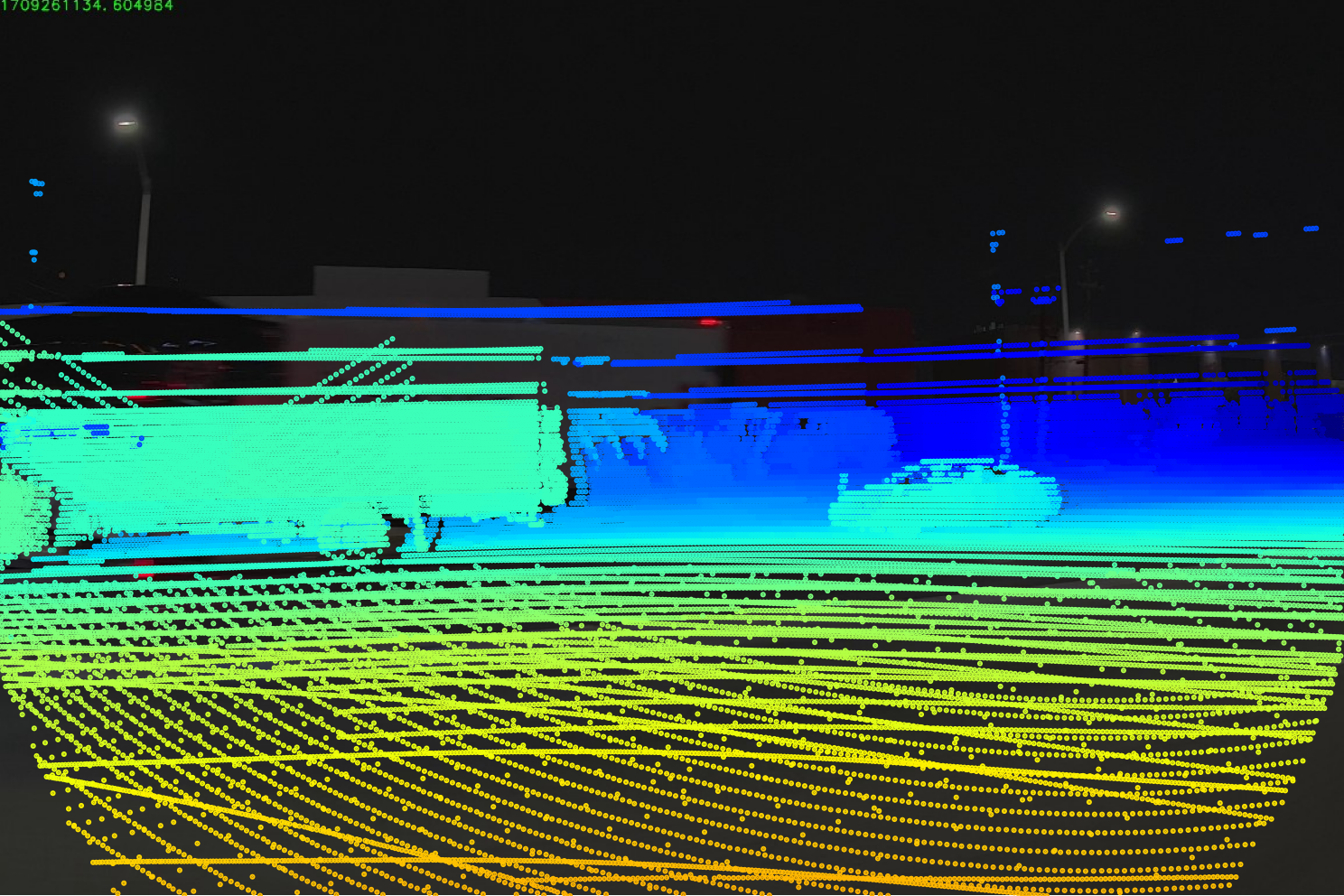}
  \centering
  % \fbox{\rule[-.5cm]{0cm}{4cm} \rule[-.5cm]{4cm}{0cm}}
  \caption{Sensor data alignment with LiDAR points projected onto the camera image and color-coded by $-log(R_i)$, where $R_i$ is per-point range value in the camera frame. Upper left: on-time camera image captured at $t=T_C$, i.e., $t^s_C=0$; Upper right: stale camera image captured at $t=T_C-0.2s$, i.e., $t^s_C=0.2s$; Lower left: on-time and synchronized LiDAR point cloud, which spatially aligns with the upper left image; Lower right: on-time LiDAR point cloud, which misaligns with the upper right image.}
  \label{fig:sensor_alignment}
  %\vspace{-0.4cm}
\end{figure}

The impact of sensor staleness is particularly significant in early/mid fusion architectures, where raw sensor data is combined directly (or processed separately through lightweight feature extractors) before the main trunk of the detection neural network~\cite{fadadu2022multi,foucard2024spotnet}. When sensors are not perfectly synchronized, this can lead to spatial misalignment of features and degraded detection performance. 
For instance in~\cref{fig:sensor_alignment}, we show real sensor data captured by our AV, where two vehicles in the image (from a port side camera) were static and ego was traveling at 30 mph from left to right. 
With the two right-side sub-figures, we simulate and demonstrate that when camera data staleness $t^s_C=0.2s$, AV's movement introduces a significant displacement during such delay and causes a large misalignment between camera image and (projected) LiDAR points. This could make it hard for the detector to output reliable detections and decide which timestamp to assign to them, which makes trajectory prediction also harder and could potentially affect safety-critical decisions.

Previous approaches to this challenge have largely focused on four strategies. The first one is attempting to perfectly synchronize sensors through hardware solutions, which is costly and often impractical in production systems, since the delay between the sensor data capture and consumption is a variable and it's distinct among sensor modalities. 
The second approach uses motion updates~\cite{della2019unified} to synchronize stale point cloud positions to the camera time, which is hard for dynamic objects in the scene and is hard the other way around (i.e., applying motion updates on camera image pixels).
The third approach is skipping model inference at the current frame. This aggregates delay for downstream components and is impractical when one particular sensor is stale for too long. 
The last approach is dropping the stale sensor data entirely and relies on the remaining sensor modalities. This approach is effective and usually applied on a model trained with sensor dropout~\cite{hwang2022cramnet,mohta2021investigating}. However, it still sacrifices valuable information from the stale sensor data. Despite these various approaches, none fully addresses the fundamental challenge of effectively utilizing stale sensor data without significant performance degradation.

Our work introduces a novel approach that enables detection models to actively learn and adapt to sensor staleness. It shows robust performance with stale sensor inputs, and its model-agnostic nature allows easy integration into existing early/mid fusion architectures.

\section{Methods}
\label{sec:methods}

\subsection{Sensor synchronization and model inputs}
\label{sec:methods_1}
We collect data with LiDARs, radars and cameras equipped on our AV fleet. LiDARs and cameras are synchronized and phase-locked on-vehicle, ensuring that (1) each LiDAR's laser beams consistently point to the same azimuth angle during rotation; (2) cameras are triggered to scan the first row when laser beams sweep to the end of their field-of-view (FoV). This approach ensures basic level of alignment between projected LiDAR point clouds and camera images, enabling good sensor fusion when all modalities are current. Details of sensor data extraction for our detection model include:

\begin{itemize}
    \item \textbf{LiDAR point cloud}: Each LiDAR rotates at 10Hz. We aggregate returns from a full sweep of individual LiDARs and merge them into a single point cloud, including 3D positions, intensity values, and per-point timestamp offsets (to be introduced in~\cref{sec:methods_2}). The LiDAR point cloud timestamp $T_L=max(T_i)$ is defined as the ``end" of each sweep (the timestamp of the latest LiDAR point).
\end{itemize}
\begin{itemize}
    \item \textbf{Camera RGB images}: The rolling shutter cameras operate at the same 10Hz frequency as LiDARs, with 5-15ms exposure time and $25\mathrm{\mu s}$ row time. The camera timestamp $T_C$ represents when the first line stops exposing.
\end{itemize}
\begin{itemize}
    \item \textbf{Radar point cloud}. To avoid interference and deal with velocity ambiguity~\cite{richards2010principles}, radars operate at different frequencies with firing time offsets. Radars are not phase-locked with LiDARs or cameras. We buffer one second of radar returns from all radars to address point sparsity, merging them into a single point cloud that includes features of 3D positions, Radar Cross Section (RCS), Signal-to-Noise Ratio (SNR), Doppler Interval, and per-point timestamp offsets (to be introduced in~\cref{sec:methods_2}). The radar point cloud timestamp $T_R=max(T_i)$ represents the latest radar point's timestamp in the buffer.
\end{itemize}

Our detection model is a perspective-view single-camera mid-fusion model trained with image data from all cameras.  For each frame, we extract LiDAR and radar point clouds in the ego local frame. We then synchronize each point's position $X_{T_i} \in \mathbb{R}^3$ to the camera timestamp $T_C$ using:
\begin{equation}{
  X_{T_C} = H_{T_C \leftarrow T_i} \cdot X_{T_i}}
  \label{eq:update_point}
\end{equation}
where $H_{T_C \leftarrow T_i} \in SE(3)$ is the transformation matrix from the individual point timestamp $T_i$ to $T_C$.
We project the updated point clouds onto the camera image and filter out points outside the image bounds. During training, we synchronize labeled 3D bounding boxes to $T_C$ using a constant velocity motion model.

\begin{figure}[t]
  \includegraphics[scale=0.22]{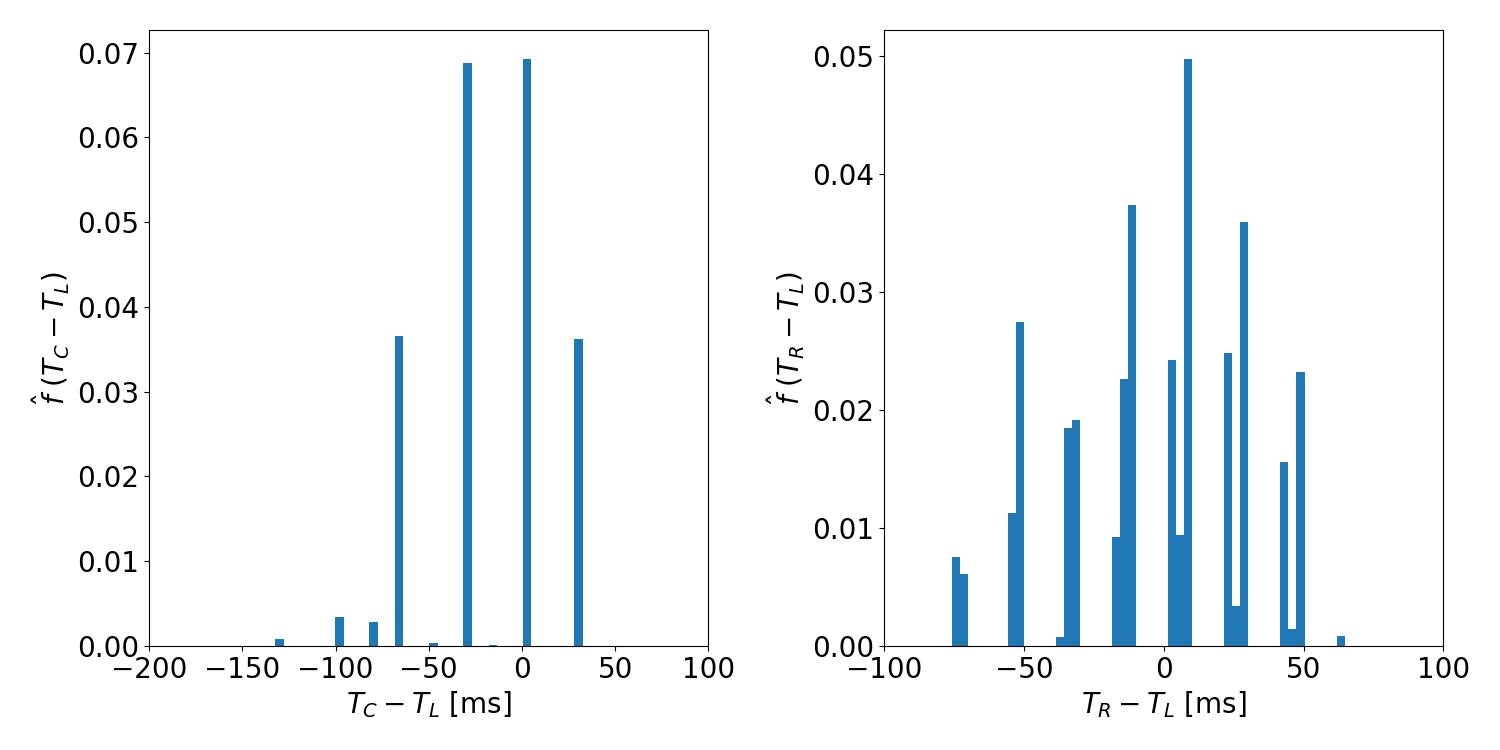}
  \centering
  % \fbox{\rule[-.5cm]{0cm}{4cm} \rule[-.5cm]{4cm}{0cm}}
  \caption{Histograms of $T_C - T_L$ (left) and $T_R - T_L$ (right) from a real on-vehicle log of 30 minutes collected at a busy scene (to stress test system latency). Y axes are normalized densities. Multiple peaks are seen since (1) sensor staleness introduces new peaks; (2) data from multiple cameras, LiDARs and radars are plotted to show the overall (mixed) distribution.}
  \label{fig:dt}
  %\vspace{-0.4cm}
\end{figure}

\subsection{Approaches dealing with sensor staleness}
\label{sec:methods_2}

We introduce two key approaches to handle sensor staleness.
First, we add a per-point timestamp offset feature for both LiDAR and radar (relative to camera): $T_C-T_i$, used as model input alongside other features (see \cref{sec:methods_1}). This provides fine-grained time information, enabling the model to learn potential sensor staleness patterns.

Second, we ``simulate" stale sensor data based on profiles from real on-vehicle logs, and apply them as data augmentation during training. In \cref{fig:dt}, unlike an ideal world with no staleness ($-0.1s<T_C - T_L<0s$), we do see out-of-distribution data where either LiDAR or camera is stale for one frame. The upper and lower bounds of $T_R - T_L$ indicate that radar-LiDAR staleness is also within one frame. 

Thus, we generate stale sensor data as follows: (1) LiDAR data and labeled 3D bounding boxes remain unchanged; (2) We calculate the perfectly synchronized camera timestamp using pure geometry:
\begin{equation}
  T_C=T_L-0.1(\theta_L-\theta_C)/(2\pi)
\end{equation}
where $\theta_L-\theta_C>0$ is the azimuth difference between the LiDAR sweep end and camera facing direction; (3) We define maximum jittering time difference $t_J^{max}$ and jitter a random timestamp $T'=T_C+\delta t$, where $\delta t$ is uniformly distributed:
\begin{equation}
  \delta t \sim \mathcal{U}(-t_J^{max}, t_J^{max})
  \label{eq:dt}
\end{equation}
We use $t_J^{max}=0.1s$ based on \cref{fig:dt}. (4) We query at $T'$ to fetch the closest camera data, allowing us to get the images one frame older (stale camera) or newer (stale LiDAR) than the synchronized image. We apply the same procedure independently for radar, except that we skip step (2), since radar is not phase-locked with camera or LiDAR. Instead, we jitter the current $T_R$ to be $T_R+\delta t$ and update the radar buffer data accordingly. Finally, we mix the stale dataset with the original dataset in training, where the stale-over-original ratio $P_S$ is a hyperparameter. 

This stale data augmentation helps the model learn both the distribution of per-point time offset feature $T_C-T_i$ and the spatial misalignment in sensor fusion when sensors are stale. We find that our method is robust and doesn't require exact replication of the time difference profile on-vehicle, only reasonable $t_J^{max}$ and $P_S$ values. 

\subsection{Sensor Fusion and Model Architecture}

\begin{figure}[t]
  \includegraphics[scale=0.18]{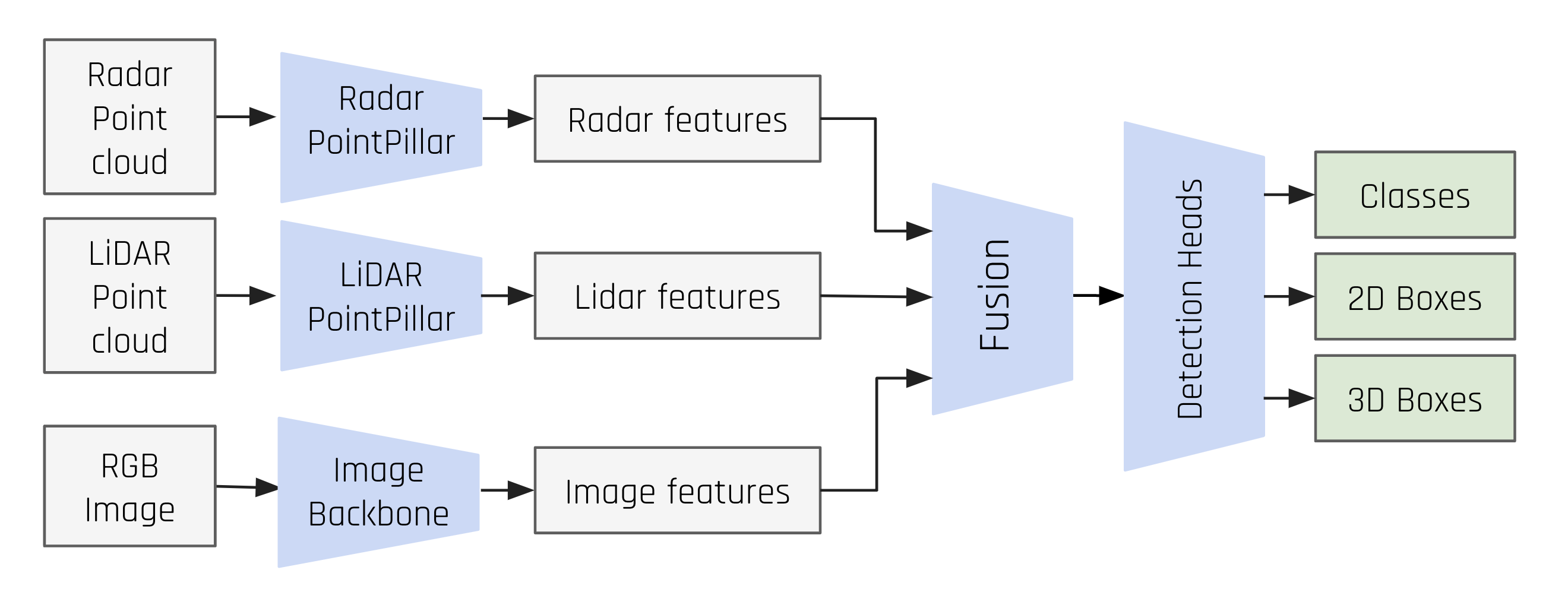}
  \centering
  % \fbox{\rule[-.5cm]{0cm}{4cm} \rule[-.5cm]{4cm}{0cm}}
  \caption{Model Architecture.}
  \label{fig:model}
  %\vspace{-0.4cm}
\end{figure}

Our approach introduces a new time offset input feature and stale sensor dataset augmentation that are model-agnostic and applicable to most early/mid sensor fusion models. Here we demonstrate it with a mid-fusion Transformer-based model in~\cref{fig:model}.
Our architecture processes each input sensor modality through an independent backbone. LiDAR and radar backbones use perspective-view PointPillar~\cite{lang2019pointpillars}, where a 
 pillar is a frustum in camera view. RGB images are processed by a YoloXPAFPN backbone~\cite{bochkovskiy2020yolov4}. Grid sizes of the LiDAR/radar pillars align with the stride 8 output dimension of the image backbone: $[B, C_{L,R,C}, H/8, W/8]$, where $B$ is the batch size, $C$ is the channel size variable for each backbone, and $H$ and $W$ are height and width of the camera image input.
 
 The output features from these backbones are fused by a dynamic fusion module~\cite{liang2022bevfusion}, followed by a Feature Pyramid Network~\cite{lin2017feature} to process and output multi-scale features (at stride 8, 16 and 32). During training, we apply feature-level sensor modality dropout with $20\%$ chance~\cite{hwang2022cramnet} by zeroing one of LiDAR/radar/Camera backbone outputs with equal probability.

We adapt DINO~\cite{zhang2022dino} as our decoder head, which uses deformable cross-attention to decode object queries to 2D boxes. Besides DINO’s original class head and 2D box head, we add a 3D box head that outputs the 3D center position, extents, yaw and velocity in the camera frame. They are supervised by human-labeled 3D bounding boxes and with an L1 loss. 
We project the labeled 3D bounding boxes onto the camera image to supervise the 2D box head with a GIoU box loss.  
The class head uses the original focal loss. We apply the Hungarian matching approach~\cite{carion2020end,zhang2022dino}, with the 2D and 3D box heads sharing the same object queries to ensure one-to-one matching between 2D and 3D outputs. 

\section{Results}

\begin{table*}[t]
\caption{Detection metrics for Cyc (bicyclists and motorcyclists), Car, and Ped (pedestrians). The highest values of each column are bolded.}
\label{table_metrics}
\centering
\setlength{\tabcolsep}{2.5pt}  % Further reduced from 4pt to 2.5pt
\small  % Reduce font size to make table more compact
\begin{tabular}{|l|l|l|ccc|ccc|ccc|}
\hline
\multirow{2}{*}{Exp. ID} & \multirow{2}{*}{Validation dataset} & \multirow{2}{*}{Model} & \multicolumn{3}{c|}{Precision} & \multicolumn{3}{c|}{Recall} & \multicolumn{3}{c|}{F1-score} \\
\cline{4-12}
 & & &  Cyc &  Car &  Ped &  Cyc &  Car &  Ped &  Cyc &  Car &  Ped \\
\hline
1a & \multirow{2}{*}{Perfectly synchronized} & baseline & 22.8\% & 40.9\% & 22.6\% & \textbf{54.0\%} & 74.5\% & \textbf{37.6\%} & \textbf{32.1\%} & 52.8\% & 28.2\% \\
1b & & w/ data augmentation & 21.6\% & 40.3\% & 22.9\% & 53.6\% & \textbf{74.6\%} & 37.5\% & 30.8\% & 52.4\% & 28.5\% \\
\hline
2a & \multirow{2}{*}{Camera staleness 100ms} & baseline & 10.2\% & 30.6\% & 7.1\% & 22.9\% & 45.6\% & 5.7\% & 14.1\% & 36.6\% & 6.3\% \\
2b & & w/ data augmentation & \textbf{23.3\%} & 40.3\% & 22.0\% & 51.3\% & 73.7\% & 34.4\% & \textbf{32.1\%} & 52.1\% & 26.8\% \\
\hline
3 & Dropout camera in inference & baseline & 22.7\% & \textbf{42.6\%} & \textbf{25.2\%} & 48.1\% & 69.6\% & 37.5\% & 30.9\% & \textbf{52.9\%} & \textbf{30.1\%} \\
\hline
\end{tabular}
\end{table*}

We evaluate our method with a proprietary dataset collected with our AV fleet. Table \ref{table_metrics} compares two models: a baseline trained with perfectly synchronized sensor data and a candidate model trained with stale sensor data augmentation (stale-over-original ratio $P_S=1.25\%$). Note that both models were trained with the per-point timestamp offset features. We cross-validated them on two validation datasets: perfectly synchronized and stale (with camera data always stale by 100ms). 

The baseline model shows significant performance degradation across all categories when evaluated on stale data (Exp. 1a vs. 2a). In contrast, the candidate model maintains robust performance (Exp. 2a vs. 2b), demonstrating that it learns to handle staleness and spatial misalignment among sensor modalities through data augmentation in training. Importantly, the candidate model achieves similar metrics on perfectly synchronized data (Exp. 1a vs. 1b), indicating that augmented stale sensor data doesn't hurt performance when $P_S$ is kept reasonably small.

\begin{figure}[t]
  \includegraphics[scale=0.27]{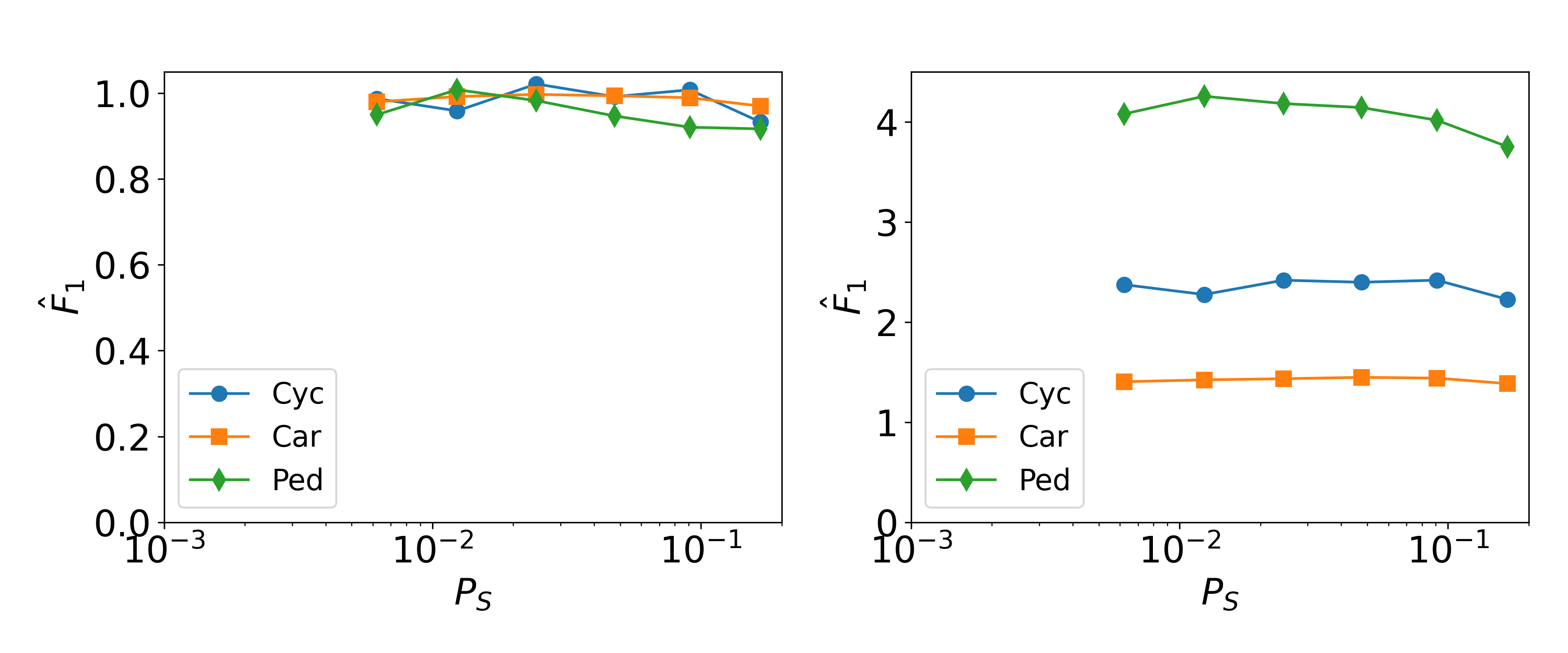}
  \centering
  % \fbox{\rule[-.5cm]{0cm}{4cm} \rule[-.5cm]{4cm}{0cm}}
  \caption{$\hat{F}_1$ (F1 score normalized by the F1 score at $P_S=0$) vs. stale-over-original ratio $P_S$ used in model training. Left: $\hat{F}_1$ evaluated with the perfectly synchronized dataset (i.e., setup 1b in \cref{table_metrics}). Right: $\hat{F}_1$ evaluated with the ``camera staleness $t^s_C=100ms$" dataset (i.e., setup 2b in \cref{table_metrics}).}
  \label{fig:f1}
  %\vspace{-0.4cm}
\end{figure}

\cref{fig:f1} shows the impact of $P_S$ (applied in training) on normalized F1 scores. A small $P_S$ (around $0.01$) provides optimal performance in both synchronized and stale conditions, showing effective learning of stale data signatures. When $P_S$ becomes too large (around $0.2$), stale data begins to contaminate the synchronized data, degrading model performance in both conditions. Interestingly, as shown in the right sub-figure, the improvement in robustness against sensor staleness is more pronounced for pedestrians than for cars across all $P_S$. This suggests that detection of smaller objects is more vulnerable to sensor data misalignment, and our approach provides the greatest benefit for these challenging cases.

For on-vehicle scenarios with excessively stale sensor data, an alternative approach is to entirely dropout this sensor input. We tested dropping out camera backbone features during inference with the baseline model (trained with sensor dropout). Results are comparable to still consuming stale data (Exp. 2b vs. 3 in \cref{table_metrics}), with the dropout approach performing worse on cyclists but better on pedestrians. However, the dropout approach faces much worse degradation when dropping out LiDAR backbone features, which is usually the dominant sensor modality in early/mid fusion models. We recommend a combined approach on-vehicle: consuming stale sensor data (with the augmentation-trained model) when staleness is below a threshold (e.g., 150ms), but applying dropout when staleness exceeds this threshold or when the sensor fails completely.

\section{Conclusion}

In this work, we presented a comprehensive approach to address the critical challenge of sensor staleness in multi-sensor fusion for autonomous vehicles. We introduced a per-point timestamp offset feature (for LiDAR and radar both relative to camera), which enables the model to learn and adapt to varying degrees of sensor staleness. Our novel data augmentation strategy simulates sensor staleness based on empirical observations from on-vehicle logs, creating realistic training scenarios without requiring any hardware modifications.
Results demonstrate that our approach significantly improves model robustness to sensor staleness while maintaining performance on perfectly synchronized data. The method is model-agnostic and can be integrated into various sensor fusion architectures with minimal modifications, and has negligible impact on system latency—making it well-suited for on-vehicle deployment.

Future work could extend our approach to incorporate rolling shutter compensation, further enhancing sensor synchronization and fusion in challenging scenarios. Additionally, we plan to evolve the current single-frame architecture into a multi-frame temporal model and evaluate our approach.

{
    \small
    \bibliographystyle{ieeenat_short}
    \bibliography{main}

\begin{thebibliography}{17}
\providecommand{\natexlab}[1]{#1}
\providecommand{\url}[1]{\texttt{#1}}
\expandafter\ifx\csname urlstyle\endcsname\relax
  \providecommand{\doi}[1]{doi: #1}\else
  \providecommand{\doi}{doi: \begingroup \urlstyle{rm}\Url}\fi

\bibitem[Bai et~al.(2022)Bai, Hu, Zhu, Huang, Chen, Fu, and Tai]{bai2022transfusion}
Xuyang Bai et~al.
\newblock Transfusion: Robust lidar-camera fusion for 3d object detection with transformers.
\newblock In \emph{Proceedings of the IEEE/CVF conference on computer vision and pattern recognition}, pages 1090--1099, 2022.

\bibitem[Bochkovskiy et~al.(2020)Bochkovskiy, Wang, and Liao]{bochkovskiy2020yolov4}
Alexey Bochkovskiy et~al.
\newblock Yolov4: Optimal speed and accuracy of object detection, 2020.

\bibitem[Carion et~al.(2020)Carion, Massa, Synnaeve, Usunier, Kirillov, and Zagoruyko]{carion2020end}
Nicolas Carion et~al.
\newblock End-to-end object detection with transformers.
\newblock In \emph{European conference on computer vision}, pages 213--229. Springer, 2020.

\bibitem[Della~Corte et~al.(2019)Della~Corte, Andreasson, Stoyanov, and Grisetti]{della2019unified}
Bartolomeo Della~Corte et~al.
\newblock Unified motion-based calibration of mobile multi-sensor platforms with time delay estimation.
\newblock \emph{IEEE Robotics and Automation Letters}, 4\penalty0 (2):\penalty0 902--909, 2019.

\bibitem[Fadadu et~al.(2022)Fadadu, Pandey, Hegde, Shi, Chou, Djuric, and Vallespi-Gonzalez]{fadadu2022multi}
Sudeep Fadadu et~al.
\newblock Multi-view fusion of sensor data for improved perception and prediction in autonomous driving.
\newblock In \emph{Proceedings of the IEEE/CVF Winter Conference on Applications of Computer Vision}, pages 2349--2357, 2022.

\bibitem[Foucard et~al.(2024)Foucard, Khanna, Shi, Liu, Shen, Ngo, and Xia]{foucard2024spotnet}
Louis Foucard et~al.
\newblock Spotnet: An image centric, lidar anchored approach to long range perception.
\newblock \emph{arXiv preprint arXiv:2405.15843}, 2024.

\bibitem[Hwang et~al.(2022)Hwang, Kretzschmar, Manela, Rafferty, Armstrong-Crews, Chen, and Anguelov]{hwang2022cramnet}
Jyh-Jing Hwang et~al.
\newblock Cramnet: Camera-radar fusion with ray-constrained cross-attention for robust 3d object detection.
\newblock In \emph{European conference on computer vision}, pages 388--405. Springer, 2022.

\bibitem[Lang et~al.(2019)Lang, Vora, Caesar, Zhou, Yang, and Beijbom]{lang2019pointpillars}
Alex~H Lang et~al.
\newblock Pointpillars: Fast encoders for object detection from point clouds.
\newblock In \emph{Proceedings of the IEEE/CVF conference on computer vision and pattern recognition}, pages 12697--12705, 2019.

\bibitem[Liang et~al.(2022)Liang, Xie, Yu, Xia, Lin, Wang, Tang, Wang, and Tang]{liang2022bevfusion}
Tingting Liang et~al.
\newblock Bevfusion: A simple and robust lidar-camera fusion framework.
\newblock \emph{Advances in Neural Information Processing Systems}, 35:\penalty0 10421--10434, 2022.

\bibitem[Lin et~al.(2017)Lin, Doll{\'a}r, Girshick, He, Hariharan, and Belongie]{lin2017feature}
Tsung-Yi Lin et~al.
\newblock Feature pyramid networks for object detection.
\newblock In \emph{Proceedings of the IEEE conference on computer vision and pattern recognition}, pages 2117--2125, 2017.

\bibitem[Liu et~al.(2021)Liu, Yu, Liu, Zhang, Qiao, Li, Tang, and Zhu]{liu2021matter}
Shaoshan Liu et~al.
\newblock The matter of time--a general and efficient system for precise sensor synchronization in robotic computing.
\newblock \emph{arXiv preprint arXiv:2103.16045}, 2021.

\bibitem[Mohta et~al.(2020)Mohta, Chou, Becker, Vallespi-Gonzalez, and Djuric]{mohta2021investigating}
A. Mohta et~al.
\newblock Investigating the effect of sensor modalities in multi-sensor detection-prediction models.
\newblock In \emph{Workshop on 'Machine Learning for Autonomous Driving' at Conference on Neural Information Processing Systems (ML4AD)}, 2020.

\bibitem[Rehder et~al.(2016)Rehder, Siegwart, and Furgale]{rehder2016general}
Joern Rehder et~al.
\newblock A general approach to spatiotemporal calibration in multisensor systems.
\newblock \emph{IEEE Transactions on Robotics}, 32\penalty0 (2):\penalty0 383--398, 2016.

\bibitem[Richards et~al.(2010)Richards, Scheer, Holm, and Melvin]{richards2010principles}
Mark~A Richards et~al.
\newblock Principles of modern radar.
\newblock 2010.

\bibitem[Yuan et~al.(2020)Yuan, Guo, and Wang]{yuan2020rggnet}
Kaiwen Yuan et~al.
\newblock Rggnet: Tolerance aware lidar-camera online calibration with geometric deep learning and generative model.
\newblock \emph{IEEE Robotics and Automation Letters}, 5\penalty0 (4):\penalty0 6956--6963, 2020.

\bibitem[Yuan et~al.(2022)Yuan, Ding, Abdelfattah, and Wang]{yuan2022licas3}
Kaiwen Yuan et~al.
\newblock Licas3: A simple lidar--camera self-supervised synchronization method.
\newblock \emph{IEEE Transactions on Robotics}, 38\penalty0 (5):\penalty0 3203--3218, 2022.

\bibitem[Zhang et~al.(2022)Zhang, Li, Liu, Zhang, Su, Zhu, Ni, and Shum]{zhang2022dino}
Hao Zhang et~al.
\newblock Dino: Detr with improved denoising anchor boxes for end-to-end object detection.
\newblock \emph{arXiv preprint arXiv:2203.03605}, 2022.

\end{thebibliography}
}


\end{document}